\documentclass{article} 
\usepackage{iclr2026_conference,times}

\usepackage{amsmath,amsfonts,bm}

\def\eqref#1{equation~\ref{#1}}

\def\1{\bm{1}}

\DeclareMathAlphabet{\mathsfit}{\encodingdefault}{\sfdefault}{m}{sl}
\SetMathAlphabet{\mathsfit}{bold}{\encodingdefault}{\sfdefault}{bx}{n}

\usepackage{booktabs}
\usepackage{hyperref}
\usepackage{url}
\usepackage{graphicx}
\graphicspath{{figures/}}

\title{InducedTFMs \\ A family of sub-quadratic tabular foundation models using induced attention}

\author{Salome Gobbi, Nicolas Atienza \& Nicola Cancedda \\
Neuralk AI
\And
Michael Arbel \\
INRIA Grenoble
}

\iclrfinalcopy 
\begin{document}

\maketitle
\lhead{Preprint} 

\begin{abstract}
Tabular foundation models (TFMs) have emerged as a serious competitor to
gradient-boosted trees, predicting on a new table in a single forward pass by
in-context learning. However, their attention is quadratic in the number of
samples and features, which caps
pretraining at small synthetic tables and forces inference on large ones to rely
on chunking or subsampling. Row-compression models reduce the feature-wise cost, but
only by discarding the cell-level representation, and remain quadratic in the
number of samples. We introduce InducedTFM, which replaces both attention stages
of a bidimensional TFM with the induced attention of the Set
Transformer~\citep{lee2019set}, preserving cell-level representations
while making its cost linear in both samples and features. Under a training
budget matched to a full-attention TabICLv2 reimplementation, InducedTFM matches
its accuracy on TabArena and on a benchmark of larger datasets, and clearly
outperforms a kernel linear-attention baseline of the same asymptotic cost. It
overtakes TabICLv2 in inference time on narrow tables, reaching a
$30\times$ speed-up at 640K rows, although row compression remains preferable on
wide tables.
\end{abstract}

\section{Introduction}

Tabular data is the dominant format for structured information in domains such
as healthcare, finance, logistics and manufacturing~\citep{9998482}. For decades,
prediction on such data has been dominated by gradient-boosted decision trees and
the AutoML ensembles built on them~\citep{chen2016xgboost, ke2017lightgbm,
prokhorenkova2018catboost, erickson2020autogluon}, all of which require a separate model to be
fitted and tuned for every new dataset. Tabular foundation models (TFMs) have
recently challenged this paradigm: pretrained once on a large distribution of
synthetic tasks, they predict on an unseen table in a single forward pass by
in-context learning over its labelled rows, without any parameter
update~\citep{hollmann2023tabpfn, hollmann2025tabpfnv2, qu2025tabicl}.

Current TFMs follow one of two architectures. \emph{Bidimensional} models, in the
line of TabPFNv2, embed every cell and alternate attention across samples and
across features at every layer, preserving a cell-level representation
throughout the network. \emph{Row-compression} models, in the line of TabICLv2,
collapse each row into a fixed-width embedding after a few feature-wise layers
and perform in-context learning over these row embeddings. In both cases,
in-context learning relies on full attention over the context rows, whose cost
is quadratic in the number of samples. Bidimensional models are in addition
quadratic in the number of features. More recently, an emerging factorized
architecture~\citep{grabocka2026ftfmfactorizedtabularfoundation} has been proposed, which keeps the cell-level
representation while replacing cell-to-cell attention with row and column latents.

This quadratic cost caps the scale at which TFMs can be trained and
served~\citep{purucker2026beyond}. Pretraining is restricted to synthetic tables
orders of magnitude smaller than many real-world datasets, and inference on
large tables must fall back on chunking, subsampling or retrieval, so the model
is never exposed to the large-scale regime in which gradient-boosted trees still
prevail. Removing this bottleneck calls for sub-quadratic attention, but most
efficient attention mechanisms developed for text or images assume a notion of
token order or locality. A TFM must instead be invariant to the order of both
rows and columns, which rules out a large part of this toolbox
(Section~\ref{related_work}).

In this work, we introduce InducedTFM, a bidimensional TFM in which both attention
stages are replaced by the induced attention of the Set
Transformer~\citep{lee2019set}. A small set of learned inducing vectors
summarises the sequence along each axis, and every token then reads back from
this summary. This keeps the model permutation invariant along both axes and
preserves the full cell-level representation, while making its cost linear in
both the number of samples and the number of features.

Our contributions are as follows:
\begin{itemize}
    \item We propose InducedTFM, to our knowledge the first TFM whose complexity is sub-quadratic along both the sample and the feature axis.
    \item Under a training budget matched to full-attention reimplementations of
    TabPFNv2 and TabICLv2, we show that InducedTFM matches TabICLv2 in predictive
    performance, both on the classification tasks of
    TabArena~\citep{erickson2025tabarenalivingbenchmarkmachine} and on a benchmark of 26 larger datasets,
    while beating a kernel linear-attention baseline of the same asymptotic cost on
    35 of the 38 TabArena datasets.
    \item We characterise inference cost across a grid of table shapes:
    InducedTFM is faster than TabPFNv2 at every shape, by up to $216\times$, and
    overtakes TabICLv2 beyond $20$K context rows on narrow tables ($\leq$ 25 features), reaching a $30\times$
    speed-up at $640$K rows.
    \item We identify the limits of this approach: the speed-up is confined to
    long and narrow tables, and a hybrid variant that additionally compresses the
    feature axis is faster still but loses accuracy, showing that the two
    mechanisms combine in cost but not in quality.
\end{itemize}

\section{Related Work}
\label{related_work}

\subsection{Deep Learning for Tabular Data}
A large body of work has tried to match gradient-boosted trees with deep
architectures, by adapting attention to sets of features~\citep{arik2021tabnet, song2019autoint, huang2020tabtransformer, somepalli2021saint, gorishniy2021revisiting}, by retrieving neighbouring training
rows~\citep{gorishniy2024tabr}, or with carefully tuned MLPs and ensembles~\citep{holzmuller2024realmlp, gorishniy2025tabm}.
Successive benchmarks have nonetheless found tuned tree ensembles to remain highly
competitive on medium-sized tables~\citep{shwartzziv2022tabular, grinsztajn2022why, mcelfresh2023when}.
Living benchmarks such as TabArena~\citep{erickson2025tabarenalivingbenchmarkmachine} now allow these
per-dataset methods to be compared directly with pretrained models.

\subsection{Tabular Foundation Models}
\label{subsec: TFMs}
Current state-of-the-art TFMs follow one of two architectures, which differ in how they treat the feature axis, and a third, factorized design has recently emerged.

\paragraph{Bidimensional attention.} TabPFNv2 \citep{hollmann2025tabpfnv2} embeds every cell and alternates attention across samples and across features at every layer, carrying the full $R \times C$ grid through the whole depth at a cost of $\mathcal{O}(C R^2 + R C^2)$ per layer. TabPFN-2.5 \citep{grinsztajn2025tabpfn25}, Mitra \citep{zhang2025mitra}, LimiX \citep{zhang2025limix} and EXAONE Tabular \citep{eo2026exaone} follow this design.

\paragraph{Row compression followed by ICL.} TabICL \citep{qu2025tabicl} first embeds each column with a Set Transformer \citep{lee2019set}, then collapses the features of each row into a few CLS tokens, so that ICL runs over one fixed-width embedding per row at a cost of $\mathcal{O}(R^2)$ independent of $C$. TabICLv2 \citep{qu2026tabiclv2}, TabPFN-3 and TabPFN-3.5 \citep{priorlabs2026tabpfn3, priorlabs2026tabpfn35} adopt this pipeline, while Causilo \citep{cho2026causilotechnicalreport}, Kumo Tabular \citep{nvidia2026kumotabular}, and 
TabFM \citep{kong2026tabfm} combine both, applying bidimensional attention over several layers before compressing rows. This trades the cell-level representation for a substantially lower cost, since most of the network then runs at a width independent of $C$.

\paragraph{Factorized cell representations.} In concurrent work, FTFM~\citep{grabocka2026ftfmfactorizedtabularfoundation} avoids this trade-off by keeping the full cell grid of bidimensional models while replacing cell-to-cell attention with a factorization. At every layer, each row and each column is pooled into a single latent, row latents attend to the training rows and column latents to each other, and every cell is updated from the latents of its own row and column. This brings the per-layer cost to $\mathcal{O}(R^2E + C^2E + RCE^2)$, the same dependence on $R$ as row-compression models while preserving cell-level representations, but it remains quadratic in the number of samples. 

\subsection{Scaling Tabular Foundation Models}

Existing work addresses the dependence on $R$ in two broad ways: scaling at inference time and subquadratic architectures. 

\paragraph{Inference-time scaling methods.} These methods reduce the context a model pretrained on small tables sees at inference. This includes subsampling or sketching the context \citep{feuer2023scaling}, retrieving the nearest neighbours of each query \citep{thomas2024locapfn}, routing queries to experts fitted on clusters of the data \citep{xu2025mixturepfn}, tuning or distilling a compact learned context \citep{feuer2024tunetables, ma2024incontext}, and learning an encoder that compresses context rows into a fixed-size memory \citep{zabergja2026taco}. These methods make large tables tractable at inference, but the underlying model is still pretrained on small synthetic tasks, so it never sees the large-scale distributions it is applied to.

\paragraph{Sub-quadratic architectures.} These methods change the model architecture so that it can be trained end-to-end on longer contexts. TabFlex \citep{zeng2025tabflex} replaces softmax attention in the original TabPFN with kernel linear attention \citep{pmlr-v119-katharopoulos20a}. Several works replace attention over samples with state-space models such as Mamba \citep{gu2023mamba}, either alone or in hybrid stacks \citep{baur2024autoregressive, wang2026somtab}, although pure state-space models were reported to underperform transformers of similar size. Moreover, as they scan their input sequentially, they are sensitive to the order of the rows. 

\paragraph{Order-agnostic efficient attention.} The state-space models above
illustrate a difficulty shared by most efficient sequence architectures: they were
designed for inputs with a natural order. Recurrent and state-space models scan
their input sequentially, and sliding-window, block-sparse or locality-based
attention~\citep{beltagy2020longformer, zaheer2020bigbird, kitaev2020reformer} assumes that nearby tokens
are related. A TFM, by contrast, should be invariant to the order of both rows and
columns. Only a subset of efficient attention mechanisms avoids this problem. Kernel and
random-feature linear attention~\citep{pmlr-v119-katharopoulos20a, choromanski2021performer}
factorise the softmax so that the cost is linear in sequence length without
referring to positions, which is the route taken by TabFlex and by our
linear-attention baseline. Bottleneck methods such as the
Perceiver~\citep{jaegle2021perceiver} and the induced set attention block of the
Set Transformer~\citep{lee2019set} instead attend through a fixed set of
learned latent or inducing vectors, which also preserves permutation invariance. Induced attention has already been
used inside TFMs, but only for column-wise embedding in TabICL~\citep{qu2025tabicl};
to our knowledge, it has not been used to make the in-context learning stage itself
sub-quadratic along both axes. Our work fills this gap: it keeps the cell-level representation of bidimensional models, like FTFM, while removing the quadratic dependence on both the number of samples and the number of features, and compares this approach directly, under a matched training budget, with the kernel linear attention used by TabFlex.

\section{InducedTFMs}
\label{InducedTFMs}
\subsection{Tabular Foundation Models}
Tabular foundation models address supervised prediction on tabular data, covering both regression and classification.
Let $D_{\mathrm{train}} := \{(x_i, y_i)\}_{i=1}^{n_{\mathrm{train}}}$ denote a labelled context, with
$x_i \in \mathcal{X} \subseteq \mathbb{R}^{C}$, where $C$ is the number of features, and $y_i \in \mathcal{Y}$ the target.
Given this context and a test input $x_{\mathrm{test}}$, the model predicts the conditional distribution
$p(y \mid x_{\mathrm{test}}, D_{\mathrm{train}})$, where $\mathcal{Y} = \{1, \dots, K\}$ for classification
and $\mathcal{Y} \subseteq \mathbb{R}$ for regression.

\subsection{Linear Complexity Architecture}
To remove the quadratic dependence on both the number of samples and the number of features described in Section \ref{subsec: TFMs}, we replace the two attention stages of a TabPFNv2-style architecture \citep{hollmann2025tabpfnv2, grinsztajn2025tabpfn25} with induced set attention blocks (ISAB) from the Set Transformer \citep{lee2019set}, which TabICL \citep{qu2025tabicl} uses for its column-wise embedding. With $N_{\mathrm{ind}}$ inducing points, the per-layer cost drops from $\mathcal{O}(RC(R+C))$ to $\mathcal{O}(N_{\mathrm{ind}}RC)$, which is linear in both the number of context samples $R$ and features $C$.

\paragraph{Embedding the table.} Write $R = R_{\text{train}} + R_{\text{test}}$
for the total number of rows and $C$ for the number of features, and let
$x_{ij} \in \mathbb{R}$ denote the entry of the table in the $i$-th row and
$j$-th column, for $1 \le i \le R$ and $1 \le j \le C$. As in TabPFNv2 \citep{hollmann2025tabpfnv2}, every entry $x_{ij}$ is
embedded independently by a single affine map shared across all cells,
$x_{ij} \mapsto x_{ij}\,w + b$, with trainable weight $w \in \mathbb{R}^{E}$
and bias $b \in \mathbb{R}^{E}$.

The target enters through a separate path. Labels are discrete, so $y_i$ is
embedded by a class embedding $\mathcal{E}_y : \{1, \dots, K\} \to
\mathbb{R}^{E}$ rather than by the cell projection, and the resulting vector is
injected twice. This is the target aware embedding (TAE). It is first added to every cell of its own row, for labelled
rows only, as done in TabICLv2 \citep{qu2026tabiclv2},
\begin{equation}
  X_{ij} = x_{ij} w + b
  \;+\; \mathbf{1}\bigl[i \le R_{\mathrm{train}}\bigr]\, \mathcal{E}_y(y_i),
  \label{eq:tae}
\end{equation}
so that a context row carries its label in all $C$ of its cells while a query
row carries none. A further column is then appended, holding
$\mathcal{E}_y(y_i)$ for context rows and a single learned vector
$y_{\mathrm{unk}} \in \mathbb{R}^{E}$, shared across all query rows, for the
rest. This gives the label a dedicated token in every row, kept separate from the feature cells, so that column-wise attention over this column can exchange label information across context and query rows directly. The embedded table is
$X \in \mathbb{R}^{R \times (C+1) \times E}$.

\begin{figure}[t]
  \centering
  \includegraphics[width=\linewidth, trim=0 0 0 0, clip]{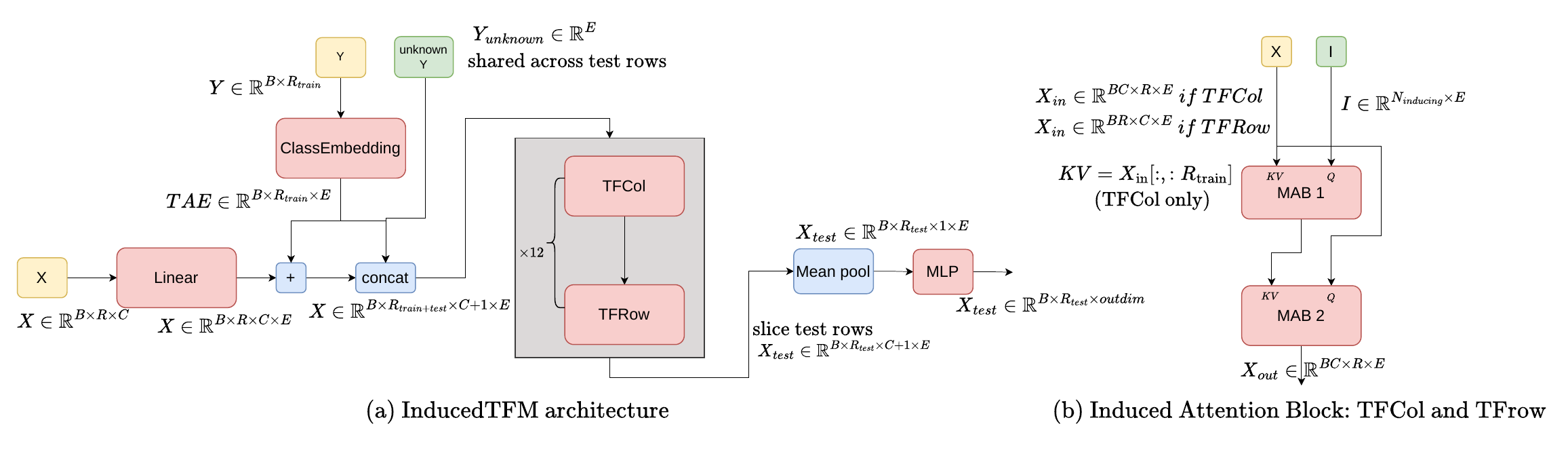}
  \caption{The InducedTFM architecture. (a) Each scalar cell is embedded by a
shared $\mathrm{Linear}(1, E)$; the target is injected twice, additively into
every cell of a train row and as an appended column in which test rows carry a
learned unknown token. The table then passes through $12$ alternating
column-wise and row-wise induced attention blocks. Test rows are mean-pooled
over features and projected to class logits by an MLP. (b) One induced
attention block: $N_{\mathrm{ind}}$ inducing vectors summarise the sequence
axis, and every token then reads back from that summary.}
  \label{fig:architecture}
\end{figure}

\paragraph{Induced attention blocks.}
An induced attention block (ISAB) replaces full self-attention over $N$ tokens
with two cross-attentions mediated by a small set of $N_{\mathrm{ind}} \ll N$ learned inducing
vectors $I \in \mathbb{R}^{N_{\mathrm{ind}} \times E}$, shared across the batch. Writing
$\mathrm{MAB}(Q, KV)$ for a pre-norm multi-head attention block with residual
connections and a feed-forward layer, the block computes
\begin{align}
  \tilde{I}          &= \mathrm{MAB}(I,\, X),          \label{eq:mab1} \\
  X_{\mathrm{out}}   &= \mathrm{MAB}(X,\, \tilde{I}),  \label{eq:mab2}
\end{align}
as illustrated in Figure~\ref{fig:architecture}(b). In the first attention the inducing
vectors act as queries: each $\tilde{I}_k$ is a softmax-weighted average of the
$N$ tokens, so the $N_{\mathrm{ind}}$ learned vectors behave as probes that each
extract one summary of the sequence. In the second, the roles are reversed and
every token queries these summaries to update its representation. Tokens thus
never attend to one another directly but exchange information in two hops through
the inducing set; composing the two attention maps
($N \times N_{\mathrm{ind}}$ and $N_{\mathrm{ind}} \times N$) yields an effective
token-to-token interaction of rank at most $N_{\mathrm{ind}}$, which is the sense
in which ISAB approximates full self-attention. Because $I$ is a fixed learned
set that is not dependent on the input or its ordering, the block remains permutation-equivariant in the tokens. Since neither step
ever forms an $N \times N$ score matrix, the cost is $\mathcal{O}(N N_{\mathrm{ind}})$ per
block rather than $\mathcal{O}(N^2)$, linear in $N$ for fixed $N_{\mathrm{ind}}$. 

\paragraph{Readout.}
After the final pass we discard the context rows, average the $C+1$ cell
embeddings of each query row, and map the result to class logits with a
two-layer MLP of hidden width $2E$.

\paragraph{InducedTFM architecture.} The full architecture of our model is shown in
Figure~\ref{fig:architecture}(a). We apply the induced attention construction along both axes.
The column-wise blocks (TFCol) treat the rows of each column as the token sequence and
summarise them with $N^{\mathrm{col}}_{\mathrm{ind}} = 128$ inducing vectors, while the
row-wise blocks (TFRow) treat the features of each row as the token sequence and
summarise them with $N^{\mathrm{row}}_{\mathrm{ind}} = 10$. In the column-wise blocks the
keys and values of \eqref{eq:mab1} are restricted to the $R_{\text{train}}$
labelled rows, so the inducing vectors summarise the context alone. Test rows read from
that summary but never contribute to it. Our model alternates 12 such TFCol/TFRow
passes, each with its own parameters, followed by the readout block producing the
class logits.

\subsection{Training}
Like other tabular foundation models, InducedTFM is not fitted to any
individual dataset. It is pretrained once on a large collection of synthetic
tabular tasks sampled from a prior $p(\mathcal{D})$, by minimising the
cross-entropy of the predicted class probabilities on the held-out rows of
each task,
\begin{equation}
  \mathcal{L} = -\,\mathbb{E}_{\mathcal{D} \sim p(\mathcal{D})}
  \sum_{i=1}^{R_{\mathrm{test}}}
  \log p_\theta\bigl(y_i \mid x_i,\, \mathcal{D}_{\mathrm{train}}\bigr).
  \label{eq:loss}
\end{equation}
At inference the weights are frozen; adapting to a new table therefore requires only a
forward pass over its context, with no gradient updates. 

\section{Experiments}
\label{experiments}

\subsection{Prior generation}
To keep the experiments reproducible we avoid a proprietary
generator and sample from the open-source TabICLv2 prior
\citep{qu2026tabiclv2}, which retains the structural causal model framework introduced in TabPFN \citep{hollmann2023tabpfn}. 
We use the released generator unchanged, except that we pre-generate
the prior and shard it to disk rather than sampling it inside the training
loop, which takes dataset generation off the critical path of the multi-GPU
runs.

\subsection{Training setup}
In order to keep the training runs tractable, we restrict them to a stage 1 training corresponding to the one done in TabICLv2 \citep{qu2026tabiclv2} and TabPFN-2.5 \citep{grinsztajn2025tabpfn25}. Each sampled task is a classification problem with up to $1024$ rows (fixed sequence length of 1024 generated rows, with 30--90\% used as context and the rest used as testing rows), up to $100$
features and $10$ classes.
We train 4 models under an identical training setup, differing only in the architecture:

\paragraph{TabPFNv2.} a reimplementation of the bidimensional full-attention architecture of TabPFNv2 from the nanoTabPFN repository~\citep{pfefferle2025nanotabpfn}, in which every layer alternates attention over the features of each row and over the rows of each column, at a cost quadratic in both the number of samples and the number of features. This is the state-of-the-art architecture that InducedTFM makes sub-quadratic: InducedTFM keeps its layout and replaces only its attention, its target injection and its output extraction.

\paragraph{TabICLv2.} a reimplementation of TabICLv2 from the nanoTabICL repository~\citep{nanotabicl}. TabICLv2 is a state-of-the-art TFM that reduces cost through a different route: after a few feature-wise layers, it compresses each row into a small number of CLS tokens, so that its ICL blocks run at a cost independent of the number of features, while remaining quadratic in the number of samples. Under our matched training budget and recipe it is less accurate than TabPFNv2, although with their respective optimised training recipes it outperforms TabPFNv2 on TabArena~\citep{qu2026tabiclv2}. It is therefore our reference for the accuracy--cost trade-off of existing efficient TFMs.

\paragraph{LinearAttn.} a baseline that replaces softmax attention with the kernel formulation of \citet{pmlr-v119-katharopoulos20a}, which TabFlex~\citep{zeng2025tabflex} used to linearise the original TabPFN. Starting from our TabICLv2 reimplementation, we replace the softmax attention in its ICL blocks with this linear attention. Since its column-wise embedding (an ISAB) is already linear in the number of rows, the resulting model is linear in the number of samples.

\paragraph{InducedTFM.} the induced-attention architecture of Section~\ref{InducedTFMs}.

All models are trained with identical optimisation settings: $100{,}000$ steps
at a global batch size of $64$, using the Muon optimiser at a learning rate of
$10^{-4}$ with gradient clipping at $10$, under a warmup--stable--decay
schedule ($1\%$ linear warmup, $90\%$ at the peak rate, then a $1-\sqrt{\cdot}$
decay to a floor of $0.1\times$ the peak). Training runs in full \texttt{fp32}
on a single node of four H100 GPUs with PyTorch DDP, the global batch being
reached by gradient accumulation over micro-batches sized to fit each model in
memory. All runs use the same seed and consume the pre-generated prior shards
in the same deterministic order, so the four architectures see identical data
in identical sequence.

\subsection{Evaluation setup}
\begin{table}[t]
\centering
{\small
\setlength{\tabcolsep}{4pt}
\begin{tabular}{@{}lcccccccc@{}}
\toprule
& & & & \multicolumn{3}{c}{Win rate (\%) $\uparrow$ vs.} & \multicolumn{2}{c}{Time (s/1K) $\downarrow$} \\
\cmidrule(lr){5-7} \cmidrule(l){8-9}
Model & Elo $\uparrow$ & Mean rank $\downarrow$ & ROC AUC $\uparrow$ & TabICLv2 & TabPFNv2 & Lin.\ attn. & Fit & Predict \\
\midrule
\multicolumn{9}{@{}l}{\textit{Our checkpoints (matched training budget)}} \\
TabPFNv2     & \textbf{1419}\,{\tiny$^{+75}_{-68}$} & \textbf{35.5} & \textbf{0.843} & \textbf{75.2} & --- & \textbf{88.6} & 3.32 & 0.323 \\
TabICLv2     & 1280\,{\tiny$^{+58}_{-65}$} & 50.1 & 0.842 & --- & \textbf{24.8} & 84.1 & 1.50 & 0.108 \\
InducedTFM & 1309\,{\tiny$^{+59}_{-67}$} & 47.0 & 0.836 & 51.8 & 21.5 & 88.0 & 2.07 & 0.186 \\
LinearAttn & 1114\,{\tiny$^{+75}_{-96}$} & 67.3 & 0.819 & 15.9 & 11.4 & --- & \textbf{1.36} & \textbf{0.077} \\
\bottomrule
\end{tabular}}
\vspace{2mm}
\caption{\textbf{TabArena results} (38 classification datasets: 30 binary, 8 multiclass). All models are evaluated in their default configuration. Elo and mean rank are computed against the full pool of 96 TabArena methods; Elo is given with $95\%$ confidence intervals. ROC AUC is averaged over folds, then over the 30 binary datasets. Win rates are measured against our implementations of TabICLv2, TabPFNv2 and LinearAttn checkpoints for baseline comparison at equal training budget; they are computed counting ties as $0.5$, averaged within each dataset and then across datasets. Fit and predict times are TabArena's median wall-clock times per 1{,}000 training samples, taken over all train/test splits.}
\label{tab:full-tabarena}
\end{table}

\paragraph{Predictive performance.} We evaluate on TabArena \citep{erickson2025tabarenalivingbenchmarkmachine} restricted to its
classification tasks, both binary and multiclass (38 datasets), and run it from the
public harness unmodified (commit \texttt{2d09631}) to ensure our results are comparable with the ones presented in TabArena's leaderboard. No hyperparameter search is performed:
each model is submitted as a single default configuration. Because these models are in-context learners, fitting does not require training, the fit call only stores the context, and all compute happens at prediction
time. Rank, win-rate and Elo are
computed by TabArena itself against its cached pool of baseline predictions. In order to cover larger dataset sizes, we have also
evaluated the models with our own protocol described in Appendix \ref{app:large-benchmark} on larger OpenML datasets we selected. 

\paragraph{Inference cost.} An accuracy benchmark only covers the shapes that those specific curated datasets
happen to take. TabArena's tasks are small enough that the asymptotic differences between
these architectures are barely exercised on it, so measuring cost there would understate
the very property a sub-quadratic model is built for. We therefore measure inference cost
separately, on synthetic tables, which lets us cover the sample--feature plane on a full grid. We sweep $9$ context sizes from
$5$K to $1.28$M rows
against feature counts from $10$ to $3200$. Wall-clock and memory depend only on tensor shapes, so both the data and the model weights are random. Each 
cell in Figure \ref{fig:speedup} is the ratio of the 2 models' inference times at the corresponding (sample, feature) size, measured as the mean of 3 forward passes (after a warmup) with $50$ query rows at batch size $1$ in \texttt{float32} on a single H100.
For inference time measurement, we use
the \emph{raw} inference path: the plain, unchunked forward which is the same computation the
model performs during training, rather than the chunked, memory-managed path used at
deployment. This is the architectural comparison: what it measures is the
attention cost itself, not the chunking machinery.

\subsection{Results}
\label{sec:results}

\begin{figure}[t]
  \centering
  \includegraphics[width=\linewidth, trim=0 0 0 0, clip]{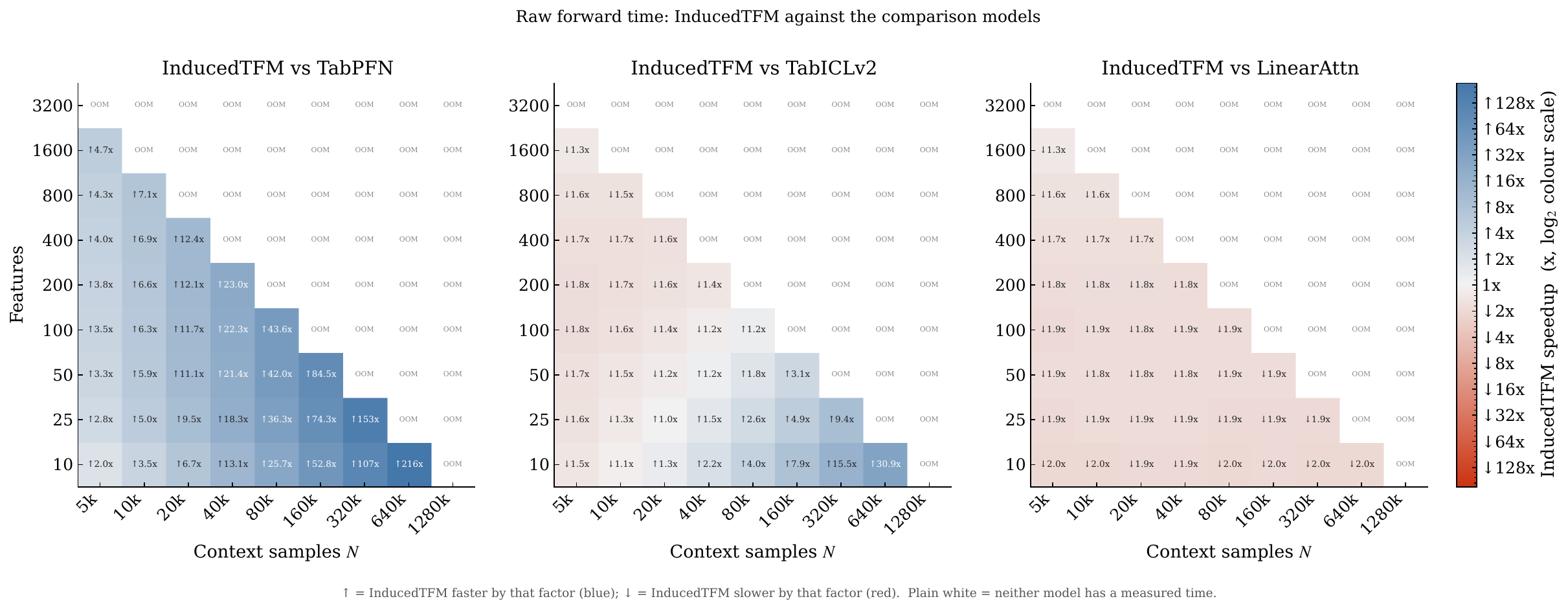}
  \caption[Raw forward-pass time of InducedTFM against TabPFNv2, TabICLv2 and linear attention]{\textbf{InducedTFM is faster than TabPFNv2 everywhere, overtakes TabICLv2 at large contexts, and trails linear attention by a constant ${\sim}2\times$ while predicting considerably better.} Ratio of raw forward-pass time between InducedTFM and TabPFNv2 (left), TabICLv2 (centre) and the linear-attention baseline (right), over a grid of context rows $R$~$\times$~features $C$. Each cell shows the speed-up ($\uparrow$, blue) or slowdown ($\downarrow$, red) of InducedTFM on a shared $\log_2$ colour scale; cells marked OOM have no measured time. Timings use the raw path: the plain, unchunked forward pass computed during training, run in eval mode without gradients, rather than the chunked, memory-managed path used at deployment, so the figure compares the attention cost of each architecture. Numbers are discussed in Section~\ref{sec:results}.}
  \label{fig:speedup}
\end{figure}

\paragraph{Predictive performance.} Table~\ref{tab:full-tabarena} reports the four
budget-matched architectures on TabArena. InducedTFM reaches an Elo of
$1309$ against $1280$ for the full-attention TabICLv2
reimplementation, $1419$ for the TabPFNv2 reimplementation and
$1114$ for the linear-attention baseline, with mean ranks of $47.0$, $50.1$,
$35.5$ and $67.3$ among the $96$ TabArena methods. The comparison with TabICLv2 is a
match rather than a win: the Elo intervals overlap, the win rate is $51.8\%$ and the
head-to-head is $23/38$ (two-sided sign test, $p=0.26$, n.s.). Raw ROC AUC over the
$30$ binary tasks shows the tie in the same way: TabICLv2 holds the better mean
($0.8423$ vs $0.8364$) and InducedTFM the better median ($0.8182$ vs $0.8169$), and
the mean gap is carried almost entirely by a single task: excluding
\texttt{Amazon\_employee\_access} the two reach $0.8433$ and $0.8425$. TabPFNv2 is the
strongest of the four: it beats both InducedTFM and TabICLv2 on $32/38$ datasets
($p<0.001$), although its mean ROC AUC ($0.8430$, or $0.8431$ without
\texttt{Amazon\_employee\_access}) is level with TabICLv2's, so its advantage is a
consistent small margin rather than a large one. The relevant separation is elsewhere in the table. The
linear-attention baseline, which replaces the softmax in the ICL blocks with the kernel
feature map of \citet{pmlr-v119-katharopoulos20a} while keeping everything else fixed,
sits roughly $170$--$200$ Elo below InducedTFM and TabICLv2 and loses $35/38$ and
$33/38$ datasets to them ($p<0.001$). Since the four models share prior shards, step
count, batch size and seed, this is attributable to the architecture alone, meaning on
tabular context, a low-rank bottleneck retains the information that a kernel feature
map of the same nominal cost discards.

\paragraph{Behaviour at larger context.} On the benchmark of larger datasets described in
Appendix~\ref{app:large-benchmark} (Table~\ref{tab:large-scale}), 26 datasets of at least
$10{,}000$ rows, the accuracy picture is unchanged: $0.8234$ for InducedTFM against $0.8228$
for TabICLv2, $10/26$ datasets won (n.s.), with TabICLv2 holding the better mean rank
($2.19$ vs $2.69$). However, the cost picture is not: InducedTFM is $1.59\times$ faster in
total inference over the 26 datasets, and the margin widens with context, from $1.32\times$
at the median dataset to $3.19\times$ on the three datasets with at least $100$K rows.
TabPFNv2 remains the most accurate model ($0.8433$ mean ROC AUC), but its lead narrows as
datasets grow. Its advantage over TabICLv2, significant on TabArena ($32/38$, $p<0.001$),
is no longer significant here ($18/26$, $p=0.076$). On the $13$ datasets above the median
size it leads TabICLv2 by only $0.003$ ROC AUC and InducedTFM by $0.009$, against $0.038$
and $0.031$ on the $13$ smaller ones. Meanwhile its cost grows fastest: TabPFNv2 is
$10.0\times$ slower than InducedTFM in total inference, and a median $27\times$ slower on
the three datasets with at least $100$K rows.

\paragraph{Inference cost across the grid.} Figure~\ref{fig:speedup} maps the ratio of raw forward time
between InducedTFM and each comparison model over the sample--feature plane: blue cells
mark where InducedTFM is faster, red cells where it is slower. All four models run out of
memory at the same shapes, so each panel compares the same $36$ cells.
Against TabPFNv2 (left), InducedTFM is faster on every cell, from $2.0\times$ at the
smallest shape ($R=5$K, $C=10$) to $216\times$ at $R=640$K, $C=10$ (median $11.4\times$).
Along the sample axis the ratio roughly doubles with every doubling of $R$, as expected
when a quadratic cost is compared with a linear one. It also grows with the number of
features, from $2.0\times$ at $C=10$ to $4.7\times$ at $C=1600$ for $R=5$K.
Against TabICLv2 (middle), the sample axis behaves as the $\mathcal{O}(RN_{\mathrm{ind}})$
cost model predicts, but with a constant factor that has to be paid first. Up to
$R=10$K, InducedTFM is slower than TabICLv2 at every shape ($1.1$--$1.8\times$),
though the penalty is small in absolute terms. Past that point the asymptotic advantage
compounds quickly: at $C=10$, InducedTFM is $1.3\times$ faster at $R=20$K, $4.0\times$ at
$80$K, $15.5\times$ at $320$K and $30.9\times$ at $640$K. The crossover moves to longer
contexts as the table widens ($R=20$K for $C\leq25$, $40$K for $C=50$, $80$K for
$C=100$), so the advantage is confined to narrow tables: InducedTFM is ahead on $15$ of
the $36$ cells, all of them at $C\leq100$.
Against the linear-attention baseline (right) there is no crossover at all. InducedTFM is
slower on all $36$ cells, and the ratio is essentially independent of the sample
dimension: at $C=10$ it stays at $1.96$--$2.0\times$ from $R=5$K to $640$K. What variation
there is runs along the feature axis instead, narrowing from $1.96\times$ at $C=10$ to
$1.32\times$ at $C=1600$. Both models therefore scale linearly in the sample dimension, and
what separates them is a constant factor of roughly $1.9\times$ (median $1.87\times$,
range $1.32$--$2.01\times$). That factor buys accuracy, however: on TabArena, InducedTFM
beats the linear attention baseline on $35/38$ datasets ($p<0.001$) and raises mean ROC AUC on the
binary tasks from $0.8194$ to $0.8364$. Dropping to pure linear attention halves the
forward cost but gives up more accuracy than the saving is worth.

\section{Limitations}
\label{sec:limitations}

\paragraph{The crossover sits above the regime where accuracy is measured.} The
sub-quadratic sample axis only pays off past roughly $20$K context rows
(Figure~\ref{fig:speedup}), and only a few TabArena tasks sit above this limit. The
accuracy results of Table~\ref{tab:full-tabarena} are therefore obtained in 
the regime where InducedTFM is the more expensive model, and  Figure~\ref{fig:speedup} shows efficiency results at shapes no established evaluation benchmark covers. 
To address this, we assemble an
evaluation suite of larger datasets that reaches into the regime where the
sub-quadratic sample axis actually pays off (Appendix~\ref{app:large-benchmark}). This
is a first step rather than a remedy: the suite is our own, and the field still
lacks an established large-scale tabular benchmark against which results in this
regime could be compared. Until one exists, efficiency and accuracy for
architectures of this kind are necessarily evaluated on different data.

\paragraph{Training was not made cheaper.} Our motivation was to lift the cap that
quadratic sample attention places on pretraining scale, and that has not yet been
demonstrated. Stage-1 tasks are $1024$ rows by up to $100$ features: a short sample axis and a wide feature axis, the one shape
in which induced sample attention buys nothing. The architecture is therefore validated
as an inference-time result on long tables, not as the world-scale training result the
motivation calls for. Establishing the latter requires a pretraining run at context
lengths where the crossover has been passed, which we have not performed.

\paragraph{Linearising the sample axis is less efficient than compressing the column axis.} 
InducedTFM keeps the TabPFNv2-style bidimensional layout, in which
attention alternates over rows and columns of a representation whose width is never
reduced. Against that, TabICLv2 compresses the feature axis to 4 CLS columns after three
passes and runs its remaining depth at a width independent of $C$. On the shapes
most real tables take (at least the ones represented in the current established benchmarks), 
that compression is worth more than an asymptotically better
sample axis.

\paragraph{Combining the two mechanisms does not combine their strengths.} We tested the
obvious response of keeping the inducing-point sample axis and adopting the feature-axis
compression as HybridTFM, described and evaluated in Appendix~\ref{app:hybrid}. The
efficiency composes: HybridTFM is the fastest of the five models across the grid and completes the large-dataset benchmark in $122$\,s against
$496$\,s for TabICLv2 and $312$\,s for InducedTFM. The accuracy does not: it reaches
$0.8010$ mean ROC AUC against $0.8228$ and $0.8234$, losing $24/26$ and $23/26$ datasets
($p<0.001$). Collapsing the feature axis to a rank-10 bottleneck and running the bulk of
the depth behind it discards information the later blocks cannot reconstruct. We report
it as an ablation rather than a competitive model, and we do not currently have an
architecture that is both fast across the whole grid and competitive on accuracy.

\paragraph{Scope.} All results are classification only, from stage-1 training at a single
seed, with one default configuration per model and no hyperparameter search or
ensembling. Regression and longer training stages are untested here, and the head-to-head sample sizes (38 and 26 datasets) leave
differences below a few points of win rate undetectable.

\section{Conclusion}
\label{sec:conclusion}

Tabular foundation models have emerged as state-of-the-art for tabular classification,
but their bidimensional attention is quadratic in both samples and features, capping the
scale at which they can be trained and served. In this work we address this by replacing full attention with the induced attention of the Set Transformer,
and we evaluate the resulting family under a training budget matched to a full-attention
reference. InducedTFM retains SOTA-level accuracy at $\mathcal{O}(R\,N_{\mathrm{ind}})$
sample cost, whereas a kernel feature-map baseline of the same nominal cost does not. The
efficiency, however, is realised only on long, narrow tables, well above the scale at
which the models are pretrained and evaluated, and the uncompressed feature axis becomes
the binding constraint on wide ones. A hybrid variant that also compresses the feature
axis is faster still but loses accuracy, indicating that the
two mechanisms compose in cost and not in quality.

\subsubsection*{Acknowledgments}
We thank the rest of the Neuralk research team, Stanislas Angebault, Aziz Fouch\'e,
Ekaterina Iakovleva, Nicolas Makaroff, Arthur Nahmias, Paul Novello and
Alexandre Pasquiou, for helpful discussions and feedback.

\bibliography{iclr2026_conference}
\bibliographystyle{iclr2026_conference}

\appendix
\section{HybridTFM: leveraging induced attention and feature-axis compression}
\label{app:hybrid}

\subsection{Architecture}

InducedTFM removes the quadratic cost in the sample dimension but leaves the
feature dimension untouched: the full $(R \times (C{+}1) \times E)$ representation is
carried through all $12$ column/row passes, so every block pays a cost proportional
to $R\,C\,N_{\mathrm{ind}}$ and holds activations proportional to $R\,C\,E$. The
feature count multiplies both, and it is never reduced. TabICLv2 instead compresses
the feature axis early: after three column/row passes it keeps only $4$ learned CLS
columns, and its $12$ ICL blocks then operate at a width independent of $C$. The
consequence is visible across the sample--feature grid of
Figure~\ref{fig:speedup}: InducedTFM is the cheaper architecture only on narrow,
long tables, and on wide ones it is slower. This is not a corner case of the grid but the regime our models
are trained in. Stage-1 tasks are $1024$ rows by up to $100$ features, so induced sample attention
buys nothing at training time. HybridTFM asks whether the two mechanisms can be
combined rather than traded off: keep the inducing-point sample axis of InducedTFM,
and adopt the feature-axis compression of TabICLv2.

\begin{figure}[t]
  \centering
  \includegraphics[width=\linewidth, trim=0 0 0 0, clip]{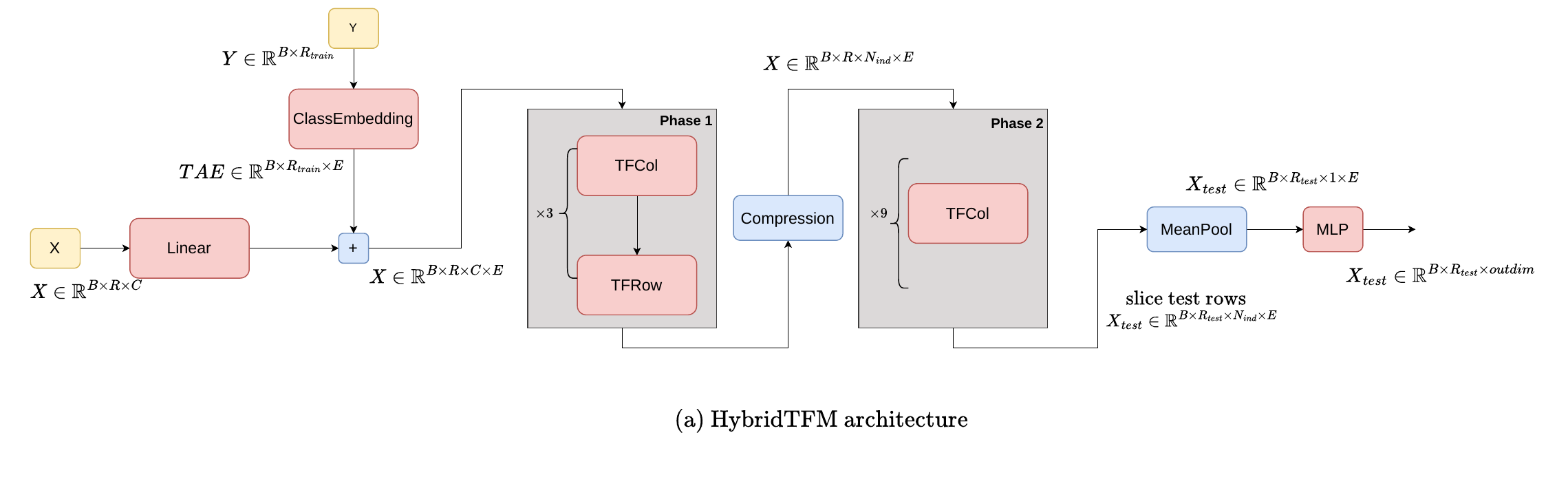}
  \caption{The HybridTFM architecture. Each scalar cell is embedded by a
shared $\mathrm{Linear}(1, E)$; the target is injected additively into
every cell of a train row. The table then passes through $3$ alternating
column-wise and row-wise induced attention blocks (128 inducing vectors in the TFCols, and 10 in the TFRows). The last row
block discards its token output and returns its $N^{\mathrm{row}}_{\mathrm{ind}}$ inducing
vectors instead, compressing each row's $C$ features into a fixed-width
summary that is independent of $C$. The remaining $9$ column-wise induced
blocks then perform in-context learning over this compressed representation.
Test rows are mean-pooled
over features and projected to class logits by an MLP. }
  \label{fig:architectureHybrid}
\end{figure}

The architecture, as seen in Figure~\ref{fig:architectureHybrid}, proceeds in two phases. 
Phase~1 alternates, for three passes, one
induced column block with one induced row block. The column block treats the $R$ rows
of a column as its token sequence and summarises them with $N^{\mathrm{col}}_{\mathrm{ind}} = 128$
inducing vectors whose keys and values are restricted to the $R_{\mathrm{train}}$
context rows; the row block treats the $C$ features of a row as its token sequence
and summarises them with $N^{\mathrm{row}}_{\mathrm{ind}} = 10$. At the third and last
row block the representation is replaced by that block's inducing-vector activations,
collapsing $(R, C, E)$ to $(R, 10, E)$: the feature axis is compressed to a fixed
width of $10$ channels. Phase~2 then runs the remaining $9$ induced
column blocks on this compressed representation, at a cost of
$\mathcal{O}(R\,N^{\mathrm{col}}_{\mathrm{ind}}\,N^{\mathrm{row}}_{\mathrm{ind}})$ per
block. Query rows are read out by mean-pooling
their $10$ channels and mapping to class logits with a two-layer MLP of hidden width
$2E$. We use $E = 128$ with $8$ heads on both axes, and inject the target additively
into every cell of a context row (the TAE path of Equation~\ref{eq:tae}) without the
appended target column.

\subsection{Experiments and results}
HybridTFM delivers the efficiency it was designed for. Across the sample--feature grid
it is the fastest of the five models (Figure~\ref{fig:speedup-annex}) and
on the large-dataset benchmark of Appendix~\ref{app:large-benchmark} it completes all
$26$ datasets in $122$\,s against $496$\,s for TabICLv2, $312$\,s for InducedTFM and
$3126$\,s for TabPFNv2, a $4.1\times$, $2.6\times$ and $25.6\times$ reduction in total
inference time. The accuracy cost,
however, is real. On TabArena, Table~\ref{tab:tabarena_annex},
HybridTFM trails InducedTFM, TabICLv2 and TabPFNv2 on every
predictive metric: it places fourth of the five architectures on Elo ($1162$
against $1309$, $1280$ and $1419$), on mean rank ($62.6$ against $47.0$, $50.1$ and
$35.5$ among the $96$ TabArena methods) and on mean ROC AUC ($0.830$ against $0.836$,
$0.842$ and $0.843$), and it wins only $24.6\%$ of comparisons against TabICLv2 and
$11.6\%$ against TabPFNv2.
It does beat the linear-attention baseline, with a $73.1\%$ win rate (Figure~\ref{fig:speedup-annex}, right).
Furthermore, on our larger benchmark whose results are in Table \ref{tab:large-scale}, HybridTFM reaches
$0.8010$ mean ROC AUC against $0.8228$ for TabICLv2, $0.8234$ for InducedTFM and
$0.8433$ for TabPFNv2, and loses head-to-head on $24/26$, $23/26$ and $26/26$ datasets
respectively ($p < 0.001$, two-sided sign test), even though it places itself above the
purely linear attention baseline it beats on $23/26$ datasets ($p < 0.001$). Compressing the feature
axis to a rank-$10$ bottleneck and running the bulk of the depth behind it therefore
discards information that the subsequent blocks cannot reconstruct. We report HybridTFM
as an ablation on this point rather than as a competitive model: it shows that the
efficiency of the two mechanisms composes cleanly while their accuracy does not.

\begin{figure}[t]
  \centering
  \includegraphics[width=\linewidth, trim=0 0 0 0, clip]{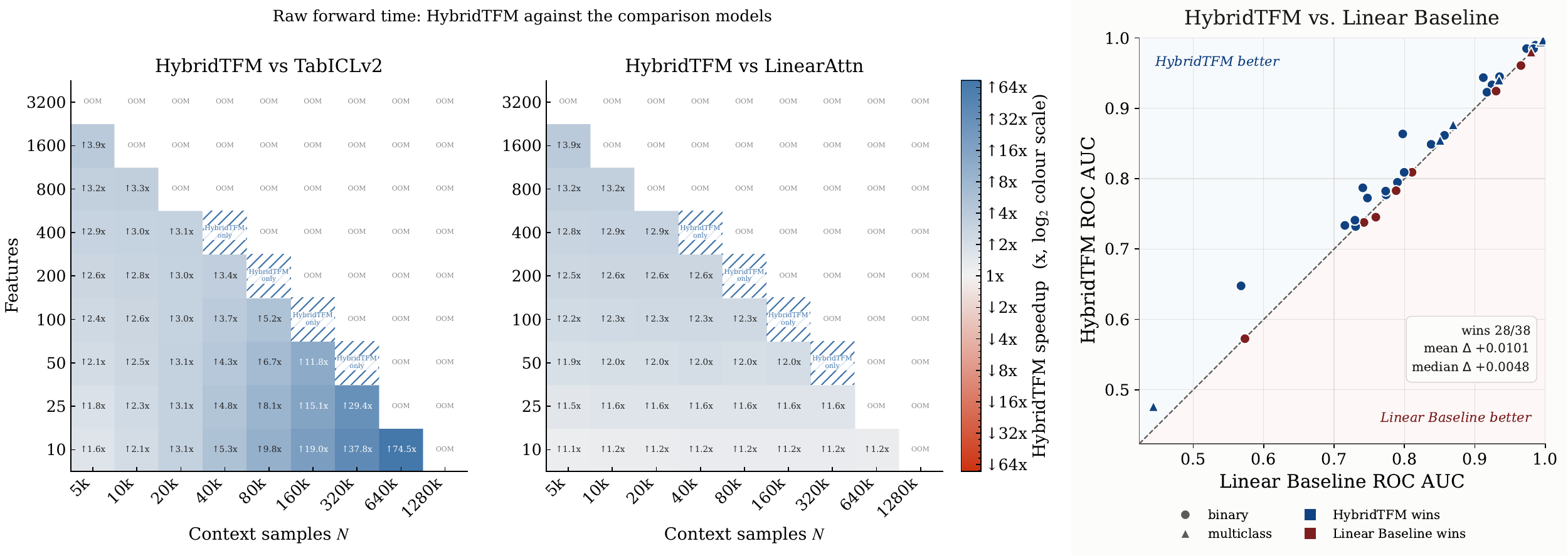}
 \caption{{\textbf{HybridTFM buys a speed advantage over quadratic
attention across the full grid, and accuracy over linear attention.}
\emph{Left and center:} per-cell ratio of raw forward-pass time over the
(context rows $R$ $\times$ features $C$) grid, annotated with the factor. $\uparrow$ marks cells where HybridTFM is faster by the stated factor (blue),
$\downarrow$ where it is slower (red); blue hatching marks cells where only HybridTFM still runs and the comparison model is out of memory; 
OOM marks cells neither model fits. Timings use
the \emph{raw} path: the plain, unchunked forward which is the same computation the
model performs during training, rather than the chunked, memory-managed path used at
deployment. This is the architectural comparison: what it measures is the
attention cost itself, not the chunking machinery. Against
TabICLv2 (left) the ratio moves with context: HybridTFM is faster across the board
but the gap grows with $R$. Against the
linear baseline (center) the ratio is essentially flat with regard to the sample dimension as they are both linear.
However, HybridTFM being also linear with regard to the feature dimension makes it advantageous in the high feature regime. 
\emph{Right:} per-dataset ROC AUC on the
38 TabArena datasets, HybridTFM against the linear attention baseline. Each point is one
dataset; points above the diagonal are
datasets HybridTFM wins, 28 of 38, with mean $\Delta$ $+0.0101$ and median
$\Delta$ $+0.0048$. }}
  \label{fig:speedup-annex}
\end{figure}

\begin{table}[t]
\centering
{\small
\setlength{\tabcolsep}{4pt}
\begin{tabular}{@{}lcccccccc@{}}
\toprule
& & & & \multicolumn{3}{c}{Win rate (\%) $\uparrow$ vs.} & \multicolumn{2}{c}{Time (s/1K) $\downarrow$} \\
\cmidrule(lr){5-7} \cmidrule(l){8-9}
Model & Elo $\uparrow$ & Mean rank $\downarrow$ & ROC AUC $\uparrow$ & TabICLv2 & TabPFNv2 & Lin.\ attn. & Fit & Predict \\
\midrule
\multicolumn{9}{@{}l}{\textit{Our checkpoints (matched training budget)}} \\
TabPFNv2     & \textbf{1419}\,{\tiny$^{+75}_{-68}$} & \textbf{35.5} & \textbf{0.843} & \textbf{75.2} & --- & \textbf{88.6} & 3.32 & 0.323 \\
TabICLv2     & 1280\,{\tiny$^{+58}_{-65}$} & 50.1 & 0.842 & --- & \textbf{24.8} & 84.1 & 1.50 & 0.108 \\
InducedTFM & 1309\,{\tiny$^{+59}_{-67}$} & 47.0 & 0.836 & 51.8 & 21.5 & 88.0 & 2.07 & 0.186 \\
HybridTFM     & 1162\,{\tiny$^{+62}_{-77}$} & 62.6 & 0.830 & 24.6 & 11.6 & 73.1 & \textbf{1.14} & 0.080 \\
LinearAttn & 1114\,{\tiny$^{+75}_{-96}$} & 67.3 & 0.819 & 15.9 & 11.4 & --- & 1.36 & \textbf{0.077} \\
\bottomrule
\end{tabular}}
\vspace{2mm}
\caption{\textbf{TabArena results} (38 classification datasets: 30 binary, 8 multiclass). All models are evaluated in their default configuration. Elo and mean rank are computed against the full pool of 96 TabArena methods; Elo is given with $95\%$ confidence intervals. ROC AUC is averaged over folds, then over the 30 binary datasets. Win rates are measured against our implementations of TabICLv2, TabPFNv2 and LinearAttn checkpoints for baseline comparison at equal training budget; they are computed counting ties as $0.5$, averaged within each dataset and then across datasets. Fit and predict times are TabArena's median wall-clock times per 1{,}000 training samples, taken over all train/test splits.}
\label{tab:tabarena_annex}
\end{table}

\section{Evaluating all models on datasets larger than TabArena}
\label{app:large-benchmark}

Since TabArena does not cover the range of dataset sizes where InducedTFM becomes interesting in terms of inference speed with regard to TabICLv2,
we have created our own benchmark composed of 26 OpenML classification datasets. Every dataset
has at least 10,000 samples (min 10,108, mean 57,154, median 45,015, max 150,000; 4--71 features,
median 11; 2--100 classes). The protocol is the following: every model is evaluated on the same 5
stratified shuffle splits per dataset (20\% test, so training contexts reach 120,000 rows). All five
checkpoints share an identical training setup so differences are attributable to architecture alone.
The results of this evaluation are found in Table~\ref{tab:large-scale}. Each metric is computed per
fold, averaged over the 5 folds to give a per-dataset score, then averaged over the 26 datasets;
ROC AUC is one-vs-one for multiclass. The $\pm$ term is the standard error across datasets, so it
measures dataset heterogeneity, not the paired difference between models. Mean rank ranks the five
models on each dataset and then averages. Inference is fit+predict wall-clock per fold summed over
the 5 folds, excluding preprocessing (identical across models) and measured with a CUDA
synchronisation. Infer.\ is its mean over datasets and Total its sum. Head-to-head cells report
datasets won by the row model, with ties excluded from the denominator (there were none;
$^{\ast}p<0.05$, $^{\ast\ast}p<0.01$, $^{\ast\ast\ast}p<0.001$, two-sided exact sign test).

\begin{table}[t]
\centering
{\small
\setlength{\tabcolsep}{4pt}
\begin{tabular}{@{}llcccccc@{}}
\toprule
 & Sample & ROC & Mean & Total & \multicolumn{3}{c}{Head-to-head} \\
\cmidrule(l){6-8}
Model & cost & AUC $\uparrow$ & rank $\downarrow$ & time (s) $\downarrow$ & Lin. & TabICLv2 & TabPFNv2 \\
\midrule
\textbf{InducedTFM} (ours) & $\mathcal{O}(RN_{\mathrm{ind}})$ & 0.8234\,{\tiny$\pm$0.029} & 2.69 & 311.9 & $26/26^{\ast\ast\ast}$ & $10/26^{\mathrm{n.s.}}$ & $1/26^{\ast\ast\ast}$ \\
TabICLv2 & $\mathcal{O}(R^2)$ & 0.8228\,{\tiny$\pm$0.033} & 2.19 & 495.5 & $25/26^{\ast\ast\ast}$ & --- & $8/26^{\mathrm{n.s.}}$ \\
TabPFNv2 & $\mathcal{O}(R^2)$ & \textbf{0.8433}\,{\tiny$\pm$0.028} & \textbf{1.35} & 3125.5 & $26/26^{\ast\ast\ast}$ & $18/26^{\mathrm{n.s.}}$ & --- \\
\textbf{HybridTFM} (ours) & $\mathcal{O}(RN_{\mathrm{ind}})$ & 0.8010\,{\tiny$\pm$0.034} & 3.92 & \textbf{122.2} & $23/26^{\ast\ast\ast}$ & $2/26^{\ast\ast\ast}$ & $0/26^{\ast\ast\ast}$ \\
LinearAttn & $\mathcal{O}(R)$ & 0.7875\,{\tiny$\pm$0.034} & 4.85 & 122.8 & --- & $1/26^{\ast\ast\ast}$ & $0/26^{\ast\ast\ast}$ \\
\bottomrule
\end{tabular}}
\vspace{2mm}
\caption{\textbf{Architecture comparison on large tabular datasets.}
26 classification datasets, each with at least $10{,}000$ samples (min $10{,}108$, mean $57{,}154$, median $45{,}015$, max $150{,}000$; 4--71 features, median 11; 2--100 classes).}
\label{tab:large-scale}
\end{table}

InducedTFM matches TabICLv2 in accuracy while costing markedly less at inference. The two are
statistically indistinguishable in terms of predictive performance: 0.8234 vs 0.8228 mean ROC AUC,
and 10/26 datasets won ($p=0.33$, n.s.), with TabICLv2 holding the better mean rank (2.19 vs 2.69),
so neither ordering is meaningful. Yet InducedTFM is $1.59\times$ faster in total inference over the
26 datasets, and the margin grows with context: $1.32\times$ at the median dataset but $3.19\times$
on the three datasets with $\geq$100K rows. Making the linear coefficient smaller is not free:
HybridTFM and the linear-attention baseline are $4\times$ faster overall ($8.3\times$ at $\geq$100K
rows) but lose to TabICLv2 on 24/26 and 25/26 datasets respectively ($p<0.001$).

TabPFNv2 is the most accurate model on this benchmark (0.8433 mean ROC AUC, mean rank 1.35), but
also by far the most expensive. It beats InducedTFM on 25/26 datasets and both linear-cost models on
26/26 ($p<0.001$), whereas its 18/26 wins over TabICLv2, the other quadratic-cost model, are not
significant ($p=0.076$). This comes at $6.3\times$ the total inference time of TabICLv2 and
$10.0\times$ that of InducedTFM; per dataset, TabPFNv2 is a median $6.9\times$ slower than InducedTFM,
rising to $27\times$ on the three datasets with $\geq$100K rows. Its accuracy advantage is moreover
concentrated on the smaller datasets: on the 13 datasets below the median size it leads InducedTFM
by 0.031 ROC AUC and TabICLv2 by 0.038, but on the 13 larger ones these gaps shrink to 0.009 and
0.003. 
In the large-context regime InducedTFM targets, TabPFNv2 therefore buys a small accuracy margin at an
order-of-magnitude higher inference cost.

\end{document}